\documentclass[10pt,twocolumn,letterpaper]{article}

\usepackage{csvsimple}
\usepackage{booktabs}  
\usepackage{cvpr}
\usepackage{times}
\usepackage{epsfig}
\usepackage{graphicx}
\usepackage{amsmath}
\usepackage{amssymb}
\usepackage{xcolor}
\usepackage{listings}
\usepackage{enumitem}

\lstdefinestyle{mystyle}{
	basicstyle=\ttfamily\small,
	breaklines=true,
	frame=single,
	columns=fullflexible,
	keepspaces=true,
	showstringspaces=false
}

\usepackage[pagebackref=true,breaklinks=true,letterpaper=true,colorlinks,bookmarks=false]{hyperref}

\cvprfinalcopy 

\def\cvprPaperID{****} 

\ifcvprfinal\fi

\begin{document}

\title{Incentivizing Neuro-symbolic Language-based Reasoning in VLMs via Reinforcement Learning}

\author{Karthic Palaniappan\\
Georgia Institute of Technology\\
{\tt\small karthik26@gatech.edu}
}

\maketitle

\begin{abstract}
There are 7,407 languages in the world. But, what about the languages that are not there in the world? Are humans so narrow minded that we don't care about the languages aliens communicate in? Aliens are humans too! In the 2016 movie Arrival, Amy Adams plays a linguist, Dr. Louise Banks who, by learning to think in an alien language (Heptapod) formed of non-sequential sentences, gains the ability to transcend time and look into the future. In this work, I aim to explore the representation and reasoning of vision-language concepts in a neuro-symbolic language, and study improvement in analytical reasoning abilities and efficiency of "thinking systems". With Qwen3-VL-2B-Instruct as base model and 4 $\times$ Nvidia H200 GPU nodes, I achieve an accuracy improvement of 3.33\% on a vision-language evaluation dataset consisting of math, science, and general knowledge questions, while reducing the reasoning tokens by 75\% over SymPy. I've documented the compute challenges faced, scaling possibilities, and the future work to improve thinking in a neuro-symbolic language in vision-language models. The training and inference setup can be found here: \href{https://github.com/i-like-bfs-and-dfs/wolfram-reasoning}{https://github.com/i-like-bfs-and-dfs/wolfram-reasoning}. 
\end{abstract}

\section{Introduction}
When solving complex problems, we think in predominantly 2 types. The first type is deduction-based deep thinking starting with a set of axioms and hypotheses, such as in International Mathematics Olympiad problems. The second type is inspiration-based thinking from a different domain, such as the development of a simulated annealing algorithm to optimize chip design inspired by annealing in metallurgy. AlphaGeometry~\cite{alphageometry} achieves SoTA results in geometric deep thinking, which was also the first automaton to win the silver medal in IMO '24 (with AlphaProof~\cite{alphaproof} in tandem). Inspiration-based thinking was solved partially by AlphaEvolve~\cite{alphaevolve}, which is able to push the boundaries of SoTA in fast matrix multiplication, efficient chip design for hardware accelerators, etc. Although these are remarkable feats, replication of human-level thinking and beyond still hasn't been achieved. 

Moreover, AlphaGeometry suffers generalizability to other mathematical domains due it it's neural engine's specificity to geometry, and AlphaEvolve requires extremely large compute due to its evolutionary nature making it infeasible to replicate under GPU constraints. 

\begin{figure}
    \centering
    \includegraphics[width=1\linewidth]{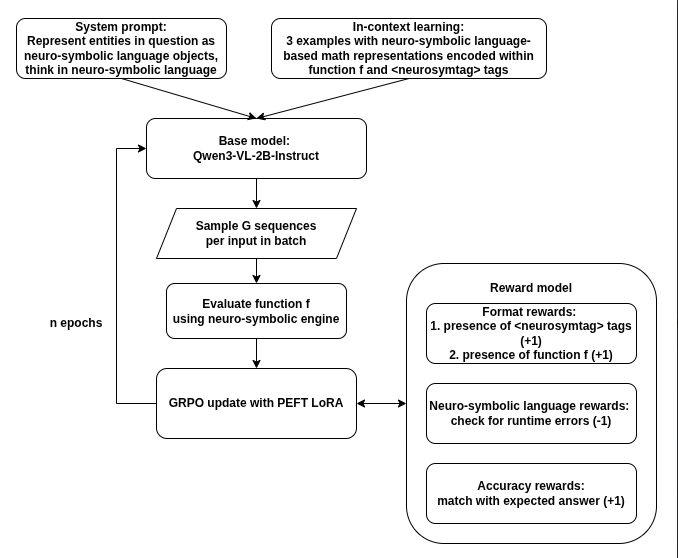}
    \caption{Overview of proposed training methodology}
    \label{fig:training_meth}
\end{figure}

To push the frontiers of generalizable problem solving of automatons across a range of mathematics concepts including solid geometry, combinatorial geometry, graph theory, etc., I've proposed a novel training methodology to incentivize vision-language foundation models to represent and think mathematics concepts in a neuro-symbolic language via reinforcement learning (Figure~\ref{fig:training_meth}). I've evaluated its mathematical VQA performance with MathVerse~\cite{mathverse}, MATH-Vision~\cite{mathvision}, and MathVista~\cite{mathvista} benchmarks. 

The neuro-symbolic language and engine is capable of representing a large set of abstract mathematics concepts (1000s), a super-set of concepts from the benchmarks mentioned above. 

In this paper, my contributions are as follows:
\begin{enumerate}[noitemsep]
\item Proposed a novel training methodology to incentivize neuro-symbolic language-based representation and thinking of mathematics concepts in Qwen3-VL-2B-Instruct model via GRPO-based RL
\item Performed a comparison of the token efficiency of the neuro-symbolic language against Python for symbolic reasoning of mathematics concepts, highlighting more human-like thinking in neuro-symbolic language
\item (WIP) Evaluated VQA performance of the trained vision-language foundation model on the MathVerse, Math-VISION, and MathVista benchmarks
\item (WIP) Evaluated the VQA generalizability and robustness of the trained vision-language foundation model to unseen mathematics concepts and input distortions 
\end{enumerate}

\section{Related Work}
AlphaGeometry trained a language model to use a deduction database as neural engine, but suffers from domain specificity of the engine and resulting restrictions in scaling. 

Generic reasoning with thinking in spoken languages has been explored by DeepSeekR1~\cite{deepseekr1}, which uses GRPO-based RL and rule-based rewards. GLM-4.5~\cite{glm45}, InternVL3.5~\cite{internvl35} are similar systems which use GRPO-based RL, CascadeRL, and accuracy-based rewards to improve performance in math benchmarks (such as AIME). Thinking in spoken languages is however very inefficient, where very large contexts are required to achieve SoTA performances (Qwen3-VL reports use of 256K context length~\cite{qwen3vl}). 

In mathematical tool use, Codex~\cite{codex}, DeepSeek, Qwen have obfuscated details of their "tool-based thinking" methodologies. MATHSENSEI~\cite{mathsensei} which explores WolframAlpha integration via in-context learning shows improvements in MATH, MMLU datasets with GPT-3, but this is early work with preliminary systems. Toolformer~\cite{toolformer} employs an SSL approach to adapt LLM with given tools (Calculator, Wikipedia, etc.), but is restricted to simple tools.

\section{Methods}
\subsection{Base model: Qwen3-VL-2B-Instruct}
For my experiments, I've used Qwen3-VL-2B-Instruct model (the model size is 4.26GB) to accomodate inference with 4x Nvidia H200 GPUs, provided by PACE at Georgia Tech. While Qwen3-VL-235B-Thinking achieves SoTA in the 3 math benchmarks, the performance of Qwen3-VL-2B-Thinking is worse by 15 to 40\%, with fairly unrestricted max. output tokens (32K tokens). This performance degrades further by 10-15\% with the Instruct version. While the the 2B Thinking version has the same model size, it produces very long CoTs in spoken language and struggles to produce any neuro-symbolic language code with in-context learning (Appendix A, Table~\ref{tab:incontext_thinking}). 

\subsection{Thinking Concepts in Neuro-symbolic Language}
\begin{figure*}
    \centering
    \includegraphics[width=1\linewidth]{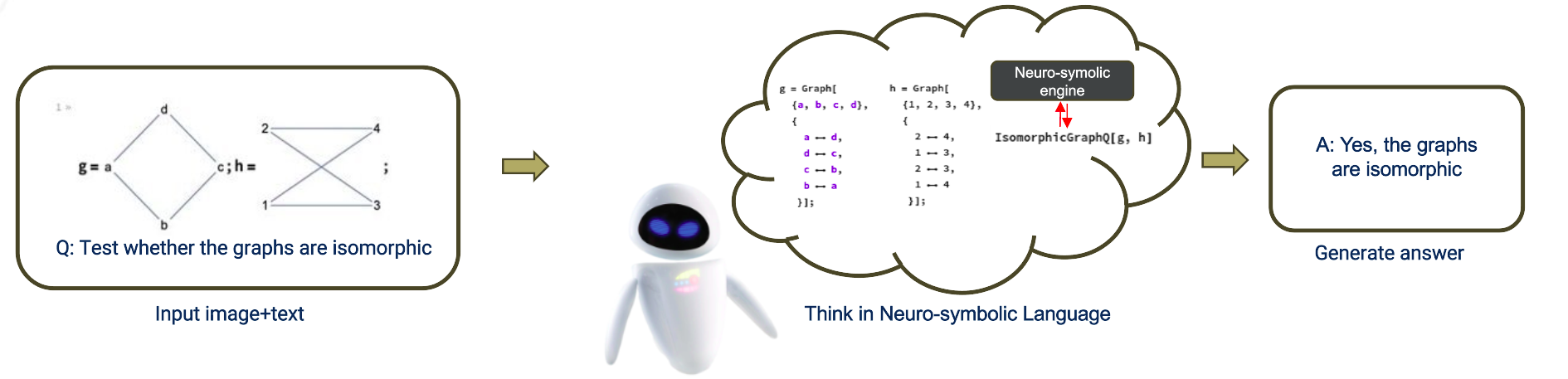}
    \caption{Thinking concepts in a neuro-symbolic language}
    \label{fig:thinkinwolfram}
\end{figure*}
Figure~\ref{fig:thinkinwolfram} shows the 3 major components to be used by tool-based VLMs in solving visual math problems via "thinking in neuro-symbolic language":
\begin{enumerate}
    \item Neuro-symbolic representation in Neuro-symbolic Language
    \item Deductions in Neuro-symbolic Engine
    \item Language model to guide representation and infer from deductions
\end{enumerate}

\subsection{Incentivizing Neuro-symbolic Language-based Reasoning via RL}
To incentivize neuro-symbolic representation in Neuro-symbolic Language, I use in-context learning~\cite{gpt3} (Figure~\ref{fig:training_meth}). More details of in-context learning are mentioned in section 4.3. To incentivize problem solving and Neuro-symbolic Engine tool calling, I use GRPO-based RL with the reward model shown in Figure~\ref{fig:training_meth}. I explore $G=10$ groups for 1 prompt (selected at random from the training set, more details in section 4.1) over 10 epochs. For low resource training, I use the PEFT LoRA~\cite{lora} to introduce low rank projections for each attention layer ($k\_proj, q\_proj, v\_proj,
o\_proj$) in the base model. This results in 6.4M trainable parameters, which is 0.3\% of the total (2.1B) in the base model. 

\section{Results}
\subsection{Dataset}
For training and evaluation, I've used the ViRL39K dataset from \cite{vlrethinker} with prompts de-duplicated (4.22\% removed), the details of which are provided in Table~\ref{tab:dataset}. The dataset contains a spread of math-related, science-related, general knowledge-related prompts, ideal to ensure performance of VLM in non-math domains. The dataset is also multi-lingual, with the English and Chinese characters appearing in both prompts and images. A 500:1 train-test split was used to keep the evaluation dataset small, due to high inference cost associated with evaluating the VLMs. 

\begin{table*}[t]
\centering

\begin{filecontents*}{data.csv}
categ,train,test,train_img,test_img
(GradeSchool) Non-Geo Math,14774,21,1.26 ± 0.75,1.00 ± 0.00
(GradeSchool) Geometric,10545,26,1.01 ± 0.14,1.04 ± 0.20
Tables/Diagrams/Charts,5258,12,1.00 ± 0.03,1.00 ± 0.00
(GradeSchool) Science,3137,3,1.00 ± 0.00,1.00 ± 0.00
Spatial Reasoning,1865,6,1.02 ± 0.16,1.17 ± 0.41
Broader STEM Topics,873,2,1.00 ± 0.00,1.00 ± 0.00
Social Science,654,2,1.00 ± 0.00,1.00 ± 0.00
Commonsense,51,0,1.00 ± 0.00,0.00 ± 0.00
Overall,37157,72,1.11 ± 0.50,1.03 ± 0.17
\end{filecontents*}
\csvreader[
    tabular={l r r l l},
    table head=\toprule
        \textbf{Category} &
        \textbf{Train Count} &
        \textbf{Test Count} &
        \textbf{Train Images per Prompt} &
        \textbf{Test Images per Prompt} \\
        \midrule,
    table foot=\bottomrule,
    late after line=\\,
    head to column names
]{data.csv}{}{
    \csvcoli & \csvcolii & \csvcoliii & \csvcoliv & \csvcolv
}
\caption{Category and train-test split details of the ViRL39K dataset. The 2nd and 3rd columns denote number of prompts in each dataset type, and the 4th and 5th columns denote the mean, std. dev. of the number of images per prompt in each category.}
\label{tab:dataset}
\end{table*}

\subsection{Baseline Results}
I evaluated the base model (with the evaluation dataset) with parameters and GPU setup detailed in Table~\ref{tab:gen-config}. A mild temperature is used to evaluate the baseline, consistent with standard practices ~\cite{stepbystep}. I used the `accelerate` huggingface library for FSDP across the GPU nodes, which uniformly distributed memory during inference across the cores. The peak memory usage was around 30\% even with the 4x Nvidia H200 GPUs, which overflowed with increase with batch size. 

\begin{table}[t]
\centering

\begin{filecontents*}{params_base.csv}
param,value
prompt,Think before you answer.
do_sample,True
temperature,0.3
max_new_tokens,4096
batch_size,1
gpu,Nvidia H200
num_gpu_procs,4
vram_per_proc,140.4GB
\end{filecontents*}
\csvreader[
    tabular={l l},   
    separator=comma,
    respect underscore=true,   
    table head=\toprule
        \textbf{Parameter} & \textbf{Value} \\
        \midrule,
    table foot=\bottomrule,
    late after line=\\
]{params_base.csv}{}{%
    \csvcoli & \csvcolii
}

\caption{Generation configuration for evaluating base model.}
\label{tab:gen-config}
\end{table}

The baseline results are shown in Table~\ref{tab:baseline_results}. The overall accuracy is 18.06\%, which shows the effect of student training with much lesser parameters (2B) compared to the teacher model (235B). The output token length reaches high values, indicating that a significant amount of resources is spent in thinking in English/Chinese languages. 

\begin{table*}[t]
\centering

\begin{filecontents*}{baseline.csv}
category,num_queries,accuracy,query_length,output_length
(GradeSchool) Geometric,26,26.92,231.50 ± 152.02,749.23 ± 1498.29
(GradeSchool) Non-Geo Math,21,14.29,300.19 ± 295.17,1704.19 ± 1959.61
(GradeSchool) Science,3,0.0,406.33 ± 267.49,10.67 ± 3.51
Broader STEM Topics,2,0.0,512.00 ± 158.39,8.50 ± 0.71
Social Science,2,0.0,442.00 ± 209.30,141.50 ± 176.07
Spatial Reasoning,6,16.67,432.50 ± 324.88,1550.00 ± 1997.37
Tables/Diagrams/Charts,12,16.67,758.83 ± 1284.48,782.58 ± 1554.00
Overall,72,18.06,377.10 ± 577.74,1031.82 ± 1679.20
\end{filecontents*}
\csvreader[
    tabular={l r r l l},
    table head=\toprule
        \textbf{Category} &
        \textbf{Count} &
        \textbf{Accuracy (\%)} &
        \textbf{PromptTokLen} &
        \textbf{OutputTokLen} \\
        \midrule,
    table foot=\bottomrule,
    late after line=\\,
    head to column names
]{baseline.csv}{}{
    \csvcoli & \csvcolii & \csvcoliii & \csvcoliv & \csvcolv
}
\caption{Evaluation results with base model. The prompt and output token lengths are averaged category-wise, with mean and std. dev. mentioned.}
\label{tab:baseline_results}
\end{table*}

\subsection{In-context Learning Results}
To get the in-context learning results, I evaluated the base model with a temperature of 1.0 (with the rest of the metrics same as that of baseline generation setup), consistent with standard practices of generation during training ~\cite{vlreasoner}. The system prompt used (shown in Listing~\ref{lst:foo}) instructs the model to perform calculations, expression simplifications, equation solving etc. in neuro-symbolic language, with 3 examples reflective of multiple math concepts, neuro-symbolic language formatting, and option matching (if required). 

The overall accuracy achieved was 6.98\%, and is reflective of few-shot performance of the base model in neuro-symbolic language-based reasoning. A significant portion of code has syntax and runtime errors, reflecting potential for improvement in neuro-symbolic  programming. A significant portion of answers don't have any code, reflecting potential for improvement in representation learning of concepts in neuro-symbolic via cold-start SFT. Since these are intermediate results, a table wasn't added, however they echo GRPO-based RL results in \ref{tab:incontext_results}.

\begin{table*}[t]
\centering
\begin{filecontents*}{incontext.csv}
category,num_queries,has_code,has_errorfree_code,accuracy,query_length,output_length
(GradeSchool) Geometric,26,92.31,53.85,11.54,710.50 ± 152.02,741.04 ± 1272.37
(GradeSchool) Non-Geo Math,21,76.19,57.14,9.52,779.19 ± 295.17,1193.76 ± 1737.28
(GradeSchool) Science,3,66.67,33.33,0.0,885.33 ± 267.49,1678.33 ± 2102.78
Broader STEM Topics,2,100.0,50.0,0.0,991.00 ± 158.39,101.50 ± 67.18
Social Science,2,100.0,50.0,0.0,921.00 ± 209.30,219.00 ± 178.19
Spatial Reasoning,6,66.67,50.0,0.0,911.50 ± 324.88,1536.33 ± 1992.92
Tables/Diagrams/Charts,12,66.67,58.33,8.33,1237.83 ± 1284.48,79.67 ± 67.90
Overall,72,80.56,54.17,8.33,856.10 ± 577.74,835.92 ± 1435.58
\end{filecontents*}
\csvreader[
    tabular={l r r r r l l},
    table head=\toprule
        \textbf{Category} &
        \textbf{Count} &
        \textbf{Code (\%)} &
        \textbf{NoErr (\%)} &
        \textbf{Accuracy (\%)} &
        \textbf{PromptTokLen} &
        \textbf{OutputTokLen} \\
        \midrule,
    table foot=\bottomrule,
    late after line=\\,
    head to column names
]{incontext.csv}{}{
    \csvcoli & \csvcolii & \csvcoliii & \csvcoliv & \csvcolv & \csvcolvi & \csvcolvii
}
\caption{Evaluation results with GRPO-based RL. Code \% denotes fraction of outputs with neuro-symbolic code. NoErr \% denotes fraction of outputs with error-free neuro-symbolic code. Accuracy denotes fraction of correct answers evaluated using the neuro-symbolic Engine. The prompt and output token lengths are averaged category-wise, with mean and std. dev. mentioned. }
\label{tab:incontext_results}
\end{table*}

\subsection{GRPO-based RL Results}
GRPO-based RL didn't improve significantly over in-context learning results due to uniform rewards over all outputs, achieving an accuracy of 8.33\%. This is however an improvement over Python (SymPy)-based tool-use (same model parameters), which resulted in an accuracy of 6\%. The results for the neuro-symbolic language tool-use are detailed in Table~\ref{tab:incontext_results}. Moreover, the updates saw only 10 training samples, the batch size of which couldn't be increased due to GPU memory overflow. 

\subsection{Ablation}
Both base and in-context learning supported a batch size 4 with a max token length of 1024 instead of 4096, but an accuracy of 0.0\% was observed in both experiments.

\section{Discussion}
In this work, I've explored an implementation of neuro-symbolic language-based reasoning of VLMs. While there is potential to improve accuracy and token length, further optimizations in inference, analysis of attention in vision component, and analysis over larger number of training datapoints are essential. 

\section{Future Work}
The experiments helped understand the importance of improving inference efficiency of VLMs. FlashAttention and vLLMs could be further explored to enable RL training with a limited set of GPUs. Multimodal low-rank projections~\cite{moka} (work recently published at NeurIPS) needs to be added to improve grounding with images, which is essential for accurate neural representation. I plan to continue working on these experiments to achieve better performance with lesser token lengths.

{\small
\bibliographystyle{ieee_fullname}
\bibliography{egbib}

\begin{thebibliography}{10}\itemsep=-1pt

\bibitem{alphaevolve}
Alexander~Novikov et. al.
\newblock Alphaevolve: A coding agent for scientific and algorithmic discovery,
  2025.

\bibitem{glm45}
Aohan~Zeng et. al.
\newblock Glm-4.5: Agentic, reasoning, and coding (arc) foundation models,
  2025.

\bibitem{mathsensei}
DEBRUP~DAS et. al.
\newblock Mathsensei: Mathematical reasoning with a tool-augmented large
  language model.
\newblock In {\em ICLR 2024 Workshop on Mathematical and Empirical
  Understanding of Foundation Models}, 2024.

\bibitem{vlrethinker}
Haozhe~Wang et. al.
\newblock {VL}-rethinker: Incentivizing self-reflection of vision-language
  models with reinforcement learning.
\newblock In {\em The Thirty-ninth Annual Conference on Neural Information
  Processing Systems}, 2025.

\bibitem{mathvista}
Pan~Lu et. al.
\newblock Mathvista: Evaluating mathematical reasoning of foundation models in
  visual contexts.
\newblock In {\em The Twelfth International Conference on Learning
  Representations}, 2024.

\bibitem{qwen3vl}
Shuai~Bai et. al.
\newblock Qwen3-vl technical report, 2025.

\bibitem{gpt3}
Tom B.~Brown et. al.
\newblock Language models are few-shot learners, 2020.

\bibitem{stepbystep}
Takeshi~Kojima et. al.
\newblock Large language models are zero-shot reasoners.
\newblock In {\em Advances in Neural Information Processing Systems}, 2022.

\bibitem{toolformer}
Timo~Schick et. al.
\newblock Toolformer: Language models can teach themselves to use tools.
\newblock In {\em Thirty-seventh Conference on Neural Information Processing
  Systems}, 2023.

\bibitem{internvl35}
Weiyun~Wang et. al.
\newblock Internvl3.5: Advancing open-source multimodal models in versatility,
  reasoning, and efficiency, 2025.

\bibitem{vlreasoner}
Yana~Wei et. al.
\newblock Open vision reasoner: Transferring linguistic cognitive behavior for
  visual reasoning.
\newblock In {\em The Thirty-ninth Annual Conference on Neural Information
  Processing Systems}, 2025.

\bibitem{deepseekr1}
Daya et.~al. Guo.
\newblock {DeepSeek-R1} incentivizes reasoning in {LLMs} through reinforcement
  learning.
\newblock {\em Nature}, 645(8081):633--638, Sept. 2025.

\bibitem{lora}
Edward~J Hu, yelong shen, Phillip Wallis, Zeyuan Allen-Zhu, Yuanzhi Li, Shean
  Wang, Lu Wang, and Weizhu Chen.
\newblock Lo{RA}: Low-rank adaptation of large language models.
\newblock In {\em International Conference on Learning Representations}, 2022.

\bibitem{alphaproof}
Thomas et.~al. Hubert.
\newblock Olympiad-level formal mathematical reasoning with reinforcement
  learning.
\newblock {\em Nature}, Nov. 2025.

\bibitem{codex}
OpenAI.
\newblock Codex performance results.
\newblock \url{https://openai.com/index/gpt-5-1-codex-max/}, 2025.
\newblock Accessed: 2025-12-10.

\bibitem{alphageometry}
Trieu H et.~al. Trinh.
\newblock Solving olympiad geometry without human demonstrations.
\newblock {\em Nature}, 625(7995):476--482, Jan. 2024.

\bibitem{mathvision}
Ke~et.~al. Wang.
\newblock Measuring multimodal mathematical reasoning with math-vision dataset.
\newblock In {\em Advances in Neural Information Processing Systems},
  volume~37, pages 95095--95169. Curran Associates, Inc., 2024.

\bibitem{moka}
Yake Wei, Yu Miao, Dongzhan Zhou, and Di Hu.
\newblock Moka: Multimodal low-rank adaptation for {MLLM}s.
\newblock In {\em The Thirty-ninth Annual Conference on Neural Information
  Processing Systems}, 2025.

\bibitem{mathverse}
Renrui et.~al. Zhang.
\newblock Mathverse: Does your multi-modal llm truly see the diagrams
  in visual math problems?
\newblock In {\em Computer Vision -- ECCV 2024}, pages 169--186, Cham, 2025.
  Springer Nature Switzerland.

\end{thebibliography}
}

\section*{Appendix}

\begin{lstlisting}[language=Python,  caption={System prompt for in-context learning of neuro-symbolic language-based reasoning},
	label={lst:foo}]
	Perform calculations and expression simplifications in a neuro-symbolic language function, assuming neuro-symbolic Engine is available to execute the code and return the answer. 
	Enclose code within <neurosymtag>f := Module[{}, ...];</neurosymtag>.
	Example 1: To find the number of circles in given image, the corresponding neuro-symbolic code is
	<neurosymtag>
	f := Module[{},
	list1 = {Circle[{0,0}, 1], Circle[{2,2}, 1]};
	Length[list1]
	];  
	</neurosymtag>
	Example 2: If a triangle ABC is given in an image with D and E bisecting AB and AC respectively, with side BC = 10cm, and the question is to find the length of DE, the corresponding neuro-symbolic code is
	<neurosymtag>
	f := Module[
	{A, B, C, D, E, DElength},
	B = {0, 0};
	C = {10, 0};
	A = {5, 5};
	D = (A + B)/2;
	E = (A + C)/2;
	DElength = EuclideanDistance[D, E];
	DElength
	];
	</neurosymtag>
	Example 3: If the question is to find the fraction of squares in an image containing 3 squares and 2 circles, with 4 options to select from (A: 3/5, B: 2/3, C: 2/5, D: 1/5), the corresponding neuro-symbolic code is
	<neurosymtag>
	f := Module[{},
	shapes = {"Square","Square","Square","Circle","Circle"};
	total = Length[shapes];
	squares = Count[shapes, "Square"];
	fraction = squares/total;
	options = <|
	"A" -> 3/5,
	"B" -> 2/3,
	"C" -> 2/5,
	"D" -> 1/5
	|>;
	SelectFirst[
	Keys[options],
	fraction == options[#] &,
	None
	]
	];
	</neurosymtag>
\end{lstlisting}

\begin{table*}[t]
\centering

\begin{filecontents*}{baseline_thinking.csv}
category,num_queries,accuracy,query_length,output_length
(GradeSchool) Geometric,26,34.62,233.50 ± 152.02,3482.15 ± 1049.88
(GradeSchool) Non-Geo Math,21,14.29,302.19 ± 295.17,3640.62 ± 934.52
(GradeSchool) Science,3,0.0,408.33 ± 267.49,4096.00 ± 0.00
Broader STEM Topics,2,50.0,514.00 ± 158.39,2904.50 ± 1685.04
Social Science,2,0.0,444.00 ± 209.30,4096.00 ± 0.00
Spatial Reasoning,6,16.67,434.50 ± 324.88,3398.67 ± 1570.33
Tables/Diagrams/Charts,12,8.33,760.83 ± 1284.48,2097.92 ± 1823.11
Overall,72,20.83,379.10 ± 577.74,3317.29 ± 1304.97
\end{filecontents*}
\csvreader[
    tabular={l r r l l},
    table head=\toprule
        \textbf{Category} &
        \textbf{Count} &
        \textbf{Accuracy (\%)} &
        \textbf{PromptTokLen} &
        \textbf{OutputTokLen} \\
        \midrule,
    table foot=\bottomrule,
    late after line=\\,
    head to column names
]{baseline_thinking.csv}{}{
    \csvcoli & \csvcolii & \csvcoliii & \csvcoliv & \csvcolv
}
\caption{Evaluation results with "thinking" model version. The prompt and output token lengths are averaged category-wise, with mean and std. dev. mentioned.}
\label{tab:baseline_thinking}
\end{table*}

\begin{table*}[t]
\centering
\begin{filecontents*}{incontext_thinking.csv}
category,num_queries,has_code,has_errorfree_code,accuracy,query_length,output_length
(GradeSchool) Geometric,26,23.08,7.69,3.85,712.50 ± 152.02,3747.88 ± 799.02
(GradeSchool) Non-Geo Math,21,28.57,9.52,0.0,781.19 ± 295.17,3606.10 ± 880.97
(GradeSchool) Science,3,0.0,0.0,0.0,887.33 ± 267.49,4096.00 ± 0.00
Broader STEM Topics,2,50.0,0.0,0.0,993.00 ± 158.39,3339.50 ± 1069.85
Social Science,2,50.0,50.0,0.0,923.00 ± 209.30,2543.00 ± 2196.27
Spatial Reasoning,6,33.33,16.67,16.67,913.50 ± 324.88,3976.83 ± 199.24
Tables/Diagrams/Charts,12,50.0,50.0,33.33,1239.83 ± 1284.48,2514.83 ± 1574.28
Overall,72,30.56,16.67,8.33,858.10 ± 577.74,3489.79 ± 1079.24
\end{filecontents*}
\csvreader[
    tabular={l r r r r l l},
    table head=\toprule
        \textbf{Category} &
        \textbf{Count} &
        \textbf{Code (\%)} &
        \textbf{NoErr (\%)} &
        \textbf{Accuracy (\%)} &
        \textbf{PromptTokLen} &
        \textbf{OutputTokLen} \\
        \midrule,
    table foot=\bottomrule,
    late after line=\\,
    head to column names
]{incontext_thinking.csv}{}{
    \csvcoli & \csvcolii & \csvcoliii & \csvcoliv & \csvcolv & \csvcolvi & \csvcolvii
}
\caption{Evaluation results with "thinking" model version and in-context learning. Code \% denotes fraction of outputs with neuro-symbolic language code. NoErr \% denotes fraction of outputs with error-free neuro-symbolic language code. Accuracy denotes fraction of correct answers evaluated using the neuro-symbolic Engine. The prompt and output token lengths are averaged category-wise, with mean and std. dev. mentioned. }
\label{tab:incontext_thinking}
\end{table*}

\end{document}